\documentclass{article}
\usepackage{spconf,amsmath,graphicx}
\usepackage{xfrac}
\usepackage[ruled,lined]{algorithm2e}
\usepackage{verbatim}
\usepackage{lipsum}

\title{The CORAL+ Algorithm for Unsupervised Domain \\ Adaptation of PLDA}
\name{Kong Aik Lee, Qiongqiong Wang, Takafumi Koshinaka}
\address{Biometrics Research Laboratories, NEC Corporation, Japan \\
\small \tt{kongaik.lee, q-wang, koshinak@nec.com}}

\begin{document}
%
\maketitle
\begin{abstract}
State-of-the-art speaker recognition systems comprise an x-vector (or i-vector) \emph{speaker embedding} front-end followed by a \emph{probabilistic linear discriminant analysis} (PLDA) backend. The effectiveness of these components relies on the availability of a large collection of labeled training data. In practice, it is common that the domains (e.g., language, demographic) in which the system is deployed differ from that we trained the system. To close the gap due to the domain mismatch, we propose an unsupervised PLDA adaptation algorithm to learn from a small amount of unlabeled in-domain data. The proposed method was inspired by a prior work on feature-based domain adaptation technique known as the \emph{correlation alignment} (CORAL). We refer to the model-based adaptation technique proposed in this paper as CORAL+. The efficacy of the proposed technique is experimentally validated on the recent NIST 2016 and 2018 Speaker Recognition Evaluation (SRE'16, SRE'18) datasets.
\end{abstract}
\begin{keywords}
Speaker recognition, domain adaptation, unsupervised, discriminant analysis
\vspace{-2ex}
\end{keywords}
\section{Introduction}
\label{sec:intro}
\vspace{-1ex}
Speaker recognition is the task of recognizing a person from his/her voice given a small amount of speech utterance from the speaker \cite{hansen2015}. Recent progresses have shown successful application of deep neural network to derive deep \emph{speaker embeddings} from speech utterances \cite{snyder2017deep, variani2014deep}. Analogous to word embeddings \cite{bengio2000,mikolov2013}, a speaker embedding is a fixed-length continuous-value vector that provides a succinct characterization of speaker’s voice rendered in a speech utterance. Similar to the classical i-vectors \cite{Dehak10frontend}, deep speaker embeddings live in a simpler Euclidean space where distance could be measured easily, compared to the much complex input patterns. Techniques like within-class covariance normalization (WCCN) \cite{Hatch2006}, linear discriminant analysis (LDA) \cite{Bishopbook}, probabilistic LDA (PLDA) \cite{Princepaper,Ioffe,kenny2010bayesian} can be applied. 

Systems comprising x-vector speaker embedding (and i-vector) followed by PLDA have shown state-of-the-art performances on speaker verification task \cite{snyder2018x}. Training an x-vector PLDA system typically requires over hundred hours of training data with speaker labels, and with the requirement that the training set must contain multiple recordings of a speaker under different settings (recording devices, transmission channels, noise, reverberation etc.). These knowledge sources contribute to the robustness of the system against such nuisance factors. The challenging problem of domain mismatch arises when a speaker recognition system is used in a different domain (e.g., different languages, demographic etc.) than that of the training data. Its performance degrades considerably.

It is impractical to re-train the system for each and every domain as the effort at collecting large labelled data sets is expensive and time consuming. A more viable solution is to adapt the already trained model using a smaller, and possibly unlabeled, set of in-domain data. Domain adaptation could be accomplished at different stages of the x-vector (or i-vector) PLDA pipeline. 
PLDA adaptation is preferable in practice since the same feature extraction and speaker embedding front-end could be used while domain adapted PLDA backbends are used to cater for the condition in each specific deployment. 

PLDA adaptation involves the adaptation of its mean vector~\footnote{Mean shift due to domain mismatch could be solved by centralizing the datasets to a common origin~\cite{Lee2017}.} and covariance matrices. In the case of unsupervised adaptation (i.e., no labels are given), the major challenge is how the adaptation could be performed on the within and between class covariance matrices given that only the total covariance matrix could be estimated directly from the in-domain data. In this paper, we show that this could be accomplished by applying similar principle as in the feature-based correlation alignment (CORAL) \cite{Sun2016} from which a pseudo-in-domain within and between class covariance matrices could be computed. We further improve the robustness by introducing additional adaptation parameter and regularization to the adaptation equation. The proposed unsupervised adaptation method is referred to as CORAL+.
\vspace{-2ex}

\section{Domain adaptation of PLDA}
\label{sec:plda_adaptation}
This section presents a brief description of \emph{probabilistic linear discriminant analysis} (PLDA) widely used in state-of-the-art speaker recognition system. We then draw attention to the domain mismatch issue and how the \emph{correlation alignment} (CORAL) \cite{Sun2016,Alam2018} technique deals with it via feature transformation.  

\subsection{Probabilistic LDA}
\label{sec:plda}
Let the vector $\phi$ be a speaker embedding (e.g., x-vector, i-vector, etc.). We assume that the vector $\phi$ is generated from a linear Gaussian model \cite{Bishopbook}, as follows \cite{Princepaper,prince2012computer}
\begin{equation}
p\left(\phi|{\bf h},{\bf x}\right) = \mathcal{N} \left( \left. \phi \right| \mu + \mathbf{Fh} + \mathbf{Gx}, \mathbf{\Sigma} \right)
\label{eq:plda}
\end{equation}
\noindent The vector $ {\bf \mu} $ represents the global mean, while $ {\bf F} $ and $ {\bf G} $ are the speaker and channel loading matrices, and the diagonal matrix $\Sigma$ models the residual variances. The variables $ {\bf h} $ and ${\bf x}$ are the latent speaker and channel variables, respectively. A PLDA model is essentially a Gaussian distribution in the speaker embedding space. This could be seen more clearly in the form of the marginal density:
\begin{equation}
p\left(\phi\right) = \mathcal{N}\left(\left. \phi \right| \mu, \mathbf{\Phi}_{\rm b}+\mathbf{\Phi}_{\rm w} \right)
\label{eq:plda_marginal}
\end{equation}
\noindent The main idea here is to account for the speaker and channel variability with a between-class and a within-class covariance matrices
\begin{equation}
 \begin{array}{l}
    \mathbf{\Phi}_{\rm b} = \mathbf{FF}^{\mathsf T} \\
    \mathbf{\Phi}_{\rm w} = \mathbf{GG}^{\mathsf T} + \mathbf{\Sigma}
 \end{array}
\end{equation}
\noindent respectively. We refer the readers to \cite{Princepaper,Ioffe,prince2012computer} for details on the model training procedure.   

In a speaker verification task, the PLDA model serves as a backend classifier. For a given pair of enrolment and test utterances, i.e, their speaker embeddings $\phi_1$ and $\phi_2$, we compute the log-likelihood ratio score 
\begin{equation}
    l\left(\phi_1, \phi_2\right) = \frac{p\left(\phi_1, \phi_2\right)}{p\left(\phi_1\right)p\left(\phi_2\right)}
\end{equation}
corresponding to the hypothesis test whether the two belong to the same or different speaker. The denominator is evaluated by substituting $\phi_1$ and $\phi_2$ in turn to \eqref{eq:plda_marginal}. The numerator is computed using 
\begin{equation}
    p\left(\phi_1, \phi_2 \right) = \mathcal{N}\left(\left. \begin{bmatrix} \phi_1 \\ \phi_2 \end{bmatrix} \right| \begin{bmatrix} \mu \\ \mu \end{bmatrix}, \begin{bmatrix} \mathbf{C} & \mathbf{\Phi}_{\rm b} \\ \mathbf{\Phi}_{\rm b} & \mathbf{C} \end{bmatrix} \right)
    \label{eq:plda_joint}
\end{equation}
where $\mathbf{C} = \mathbf{\Phi}_{\rm b} + \mathbf{\Phi}_{\rm w}$ is the total covariance matrix. The assumption is that the unseen data follow the same distribution as given by the within and between classes covariance matrices derived from the training set (i.e., the dataset we used to train the PLDA). Problem arises when the training set was drawn from a domain (out-of-domain) different from that of the enrollment and test utterances (in-domain).

\subsection{Correlation Alignment}
\label{sec:coral}

\emph{Correlation alignment} (CORAL) \cite{Sun2016} aims to align the second-order statistics, i.e., covariance matrices, of the out-of-domain (OOD) features to match the in-domain (InD) features. No class (i.e., speaker) label is used and therefore it belongs to the class of unsupervised adaptation techniques. The algorithm consists of two steps, namely, whitening followed by re-coloring. Let $\mathbf{C}_{\rm o}$ and $\mathbf{C}_{\rm I}$ be the covariance matrices of the OOD and InD data, respectively. Denote $\phi$ as a OOD vector, domain adaptation is performed by first whitening and then re-coloring, as follows

\begin{equation}
    \phi^{'} = \mathbf{C}_{\rm I}^{\frac{1}{2}} \mathbf{C}_{\rm o}^{-\frac{1}{2}} \phi    
\end{equation}
where
\begin{equation}
   \mathbf{C}_{\rm o}^{-\frac{1}{2}} = \mathbf{Q}_{\rm o} \mathbf{\Lambda}_{\rm o}^{-\frac{1}{2}}     \mathbf{Q}_{\rm o}^{\mathsf T}
   \notag
\end{equation}
whitens the input vector, and
\begin{equation}
   \mathbf{C}_{\rm I}^{\frac{1}{2}} = \mathbf{Q}_{\rm I} \mathbf{\Lambda}_{\rm I}^{\frac{1}{2}}     \mathbf{Q}_{\rm I}^{\mathsf T}
   \notag
\end{equation}
does the re-coloring. Here, $\mathbf{Q}$ and $\mathbf{\Lambda}$ are the eigenvectors and eigenvalues pertaining to the covariance matrices\footnote{The whitening and re-coloring procedures are better known as the zero-phase component analysis (ZCA) transformation \cite{Kessy2018}. As opposed to principal component analysis (PCA) and Cholesky whitening (and re-coloring), ZCA preserves the maximal similarity of the transformed feature to the original space.}. 
Such simpler and ``frustratingly easy'' approach \cite{Alam2018} has shown to outperform a more complicated non-linear transformation reported in \cite{Lin2018}. In \cite{Alam2018}, CORAL is performed on the OOD x-vectors (or i-vectors) embeddings, and the transformed vectors (pseudo in-domain) are used to re-train the PLDA. Note that speaker labels of the OOD training data remain the same.     

\section{The CORAL+ Algoritm}
\label{sec:coral+}
CORAL is a feature-based domain adaptation technique \cite{Sun2016}. We propose integrating CORAL to PLDA leading to a model-based domain adaptation.

\subsection{Domain adaptation}
It is commonly known that a linear transformation on a normally distributed vector leads to an equivalent transformation on the mean vector and covariance matrix of its density function. Let 
$\mathbf{A} =\mathbf{C}_{\rm I}^{1/2} \mathbf{C}_{\rm o}^{-1/2}$ be the transformation matrix and $\phi^{'} = \mathbf{A}^{\mathsf{T}}\phi$ the transformed vector. The covariance matrix of the pseudo in-domain vector $\phi^{'}$ is given by
\begin{equation}
 \begin{aligned}
   \mathbf{C}_{\rm o}^{'} = \mathbf{A}^{\mathsf{T}} \mathbf{C}_{\rm o} \mathbf{A} 
   =  \mathbf{A}^{\mathsf{T}} \mathbf{\Phi}_{\rm w,o} \mathbf{A}  + \mathbf{A}^{\mathsf{T}} \mathbf{\Phi}_{\rm b,o} \mathbf{A} 
 \end{aligned}
\end{equation}
Here, we have considered a PLDA trained on OOD data with a total covariance matrix $\mathbf{C}_{\rm o} = \mathbf{\Phi}_{\rm w,o}  + \mathbf{\Phi}_{\rm b,o}$ given by the sum of within and between class covariance matrices, as noted in Section \ref{sec:plda}. The above equation shows that training a PLDA on the transformed vectors $\phi^{'}$, as proposed in \cite{Alam2018}, is equivalent to transforming the within-class, between-class, and total covariance matrices of a PLDA trained on OOD data.

\subsection{Model-level adaptation}
Instead of replacing the covariance matrices in an OOD PLDA with pseudo in-domain matrices, model-level adaptation allows us to consider their interpolation
\begin{equation}
    \begin{aligned}
         \mathbf{\Phi}^{+}_{\rm b} =& (1-\beta)\mathbf{\Phi}_{\rm b,o} + \beta \mathbf{A}^{\mathsf{T}} \mathbf{\Phi}_{\rm b,o} \mathbf{A}   \\
         \mathbf{\Phi}^{+}_{\rm w} =& (1-\gamma)\mathbf{\Phi}_{\rm w,o} + \gamma \mathbf{A}^{\mathsf{T}} \mathbf{\Phi}_{\rm w,o} \mathbf{A}  
    \end{aligned}
    \notag
\end{equation}
where $\{\beta, \gamma\}$ are the adaptation parameters constrained to lie between zero and one. Notice that the first term on the right-hand-side of the equations is the OOD between/within covariance matrix while the second term is the pseudo-in-domain covariance matrix. For clarity, we further simplify the adaptation equations, as follows
\begin{equation}
    \begin{aligned}
         \mathbf{\Phi}^{+}_{\rm b} =& \mathbf{\Phi}_{\rm b,o} + \beta \left(\mathbf{A}^{\mathsf{T}} \mathbf{\Phi}_{\rm b,o} \mathbf{A} - \mathbf{\Phi}_{\rm b,o} \right)   \\
         \mathbf{\Phi}^{+}_{\rm w} =& \mathbf{\Phi}_{\rm w,o} + \gamma \left(\mathbf{A}^{\mathsf{T}} \mathbf{\Phi}_{\rm w,o} \mathbf{A} - \mathbf{\Phi}_{\rm w,o} \right)  
    \end{aligned}
    \label{eq:model_adapt}
\end{equation}
The second term on the right-hand-side of the equations represents the new information seen in the in-domain data to be added to the PLDA model. 

\begin{figure}[t!]
    \caption{The effects of regularization. Elements with negative variances are removed automatically.}
	\centering
	\includegraphics[width=2.8in]{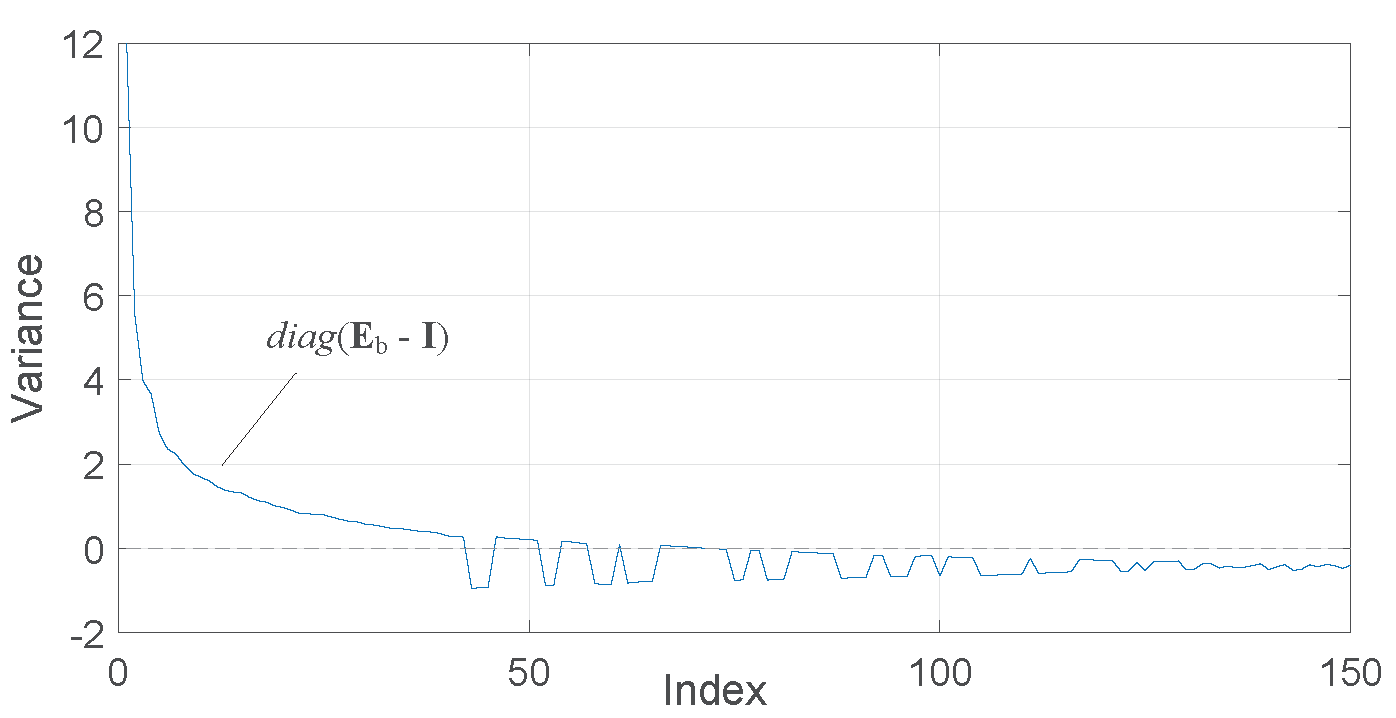}
	\label{fig:regul}
\end{figure}

\subsection{Regularized adaptation}
The central idea of domain adaptation is to propagate the uncertainty seen in the in-domain data to the PLDA model. The adaptation equations in \eqref{eq:model_adapt}, do not guarantee that the variances, and therefore the uncertainty, increase. In this section, we achieve this goal in the transform space where both the OOD and pseudo-in-domain matrices are simultaneously diagonalized.

Let $\mathbf{B}$ be an orthogonal matrix such that $\mathbf{B}^{\mathsf{T}} \mathbf{\Phi} \mathbf{B} = \mathbf{I}$ and $\mathbf{B}^{\mathsf{T}} \left(\mathbf{A}^{\mathsf{T}} \mathbf{\Phi} \mathbf{A}  \right) \mathbf{B} = \mathbf{E}$, where $\mathbf{E}$ is a diagonal matrix. This procedure is referred to as \emph{simultaneous diagonalization}. The transformation matrix $\mathbf{B}$ is obtained by performing twice the \emph{eigenvalue decomposition} (EVD) on the matrix $\mathbf{\Phi}$ and then $\mathbf{A}^{\mathsf{T}} \mathbf{\Phi} \mathbf{A}$ after the first transformation has been applied. The procedure is illustrated in Algorithm \ref{alg:coral+}.

By applying the simultaneous diagonalization on \eqref{eq:model_adapt}, the following adaptation could be obtained:    
\begin{equation}
    \begin{aligned}
         \mathbf{\Phi}^{+}_{\rm b} =& \mathbf{\Phi}_{\rm b,o} + \beta \mathbf{B}_{\rm b}^{-\mathsf{T}} \left(\mathbf{E}_{\rm b}  - \mathbf{I} \right) \mathbf{B}_{\rm b}^{-1}   \\
         \mathbf{\Phi}^{+}_{\rm w} =& \mathbf{\Phi}_{\rm w,o} + \gamma \mathbf{B}_{\rm w}^{-\mathsf{T}} \left(\mathbf{E}_{\rm w} - \mathbf{I} \right) \mathbf{B}_{\rm w}^{-1}   
    \end{aligned}
    \label{eq:coral+_wo_reg}
\end{equation}
As before, the between and within class covariance matrices are adapted separately. Notice that the term $\left(\mathbf{E} - \mathbf{I} \right)$ will ends up with negative variances if any diagonal elements of $\mathbf{E}$ is less than one. We propose the following regularized adaptation:       
\begin{equation}
    \begin{aligned}
         \mathbf{\Phi}^{+}_{\rm b} =& \mathbf{\Phi}_{\rm b,o} + \beta \mathbf{B}_{\rm b}^{-\mathsf{T}} max\left(\mathbf{E}_{\rm b}  - \mathbf{I} \right) \mathbf{B}_{\rm b}^{-1}   \\
         \mathbf{\Phi}^{+}_{\rm w} =& \mathbf{\Phi}_{\rm w,o} + \gamma \mathbf{B}_{\rm w}^{-\mathsf{T}} max\left(\mathbf{E}_{\rm w} - \mathbf{I} \right) \mathbf{B}_{\rm w}^{-1}   
    \end{aligned}
    \label{eq:coral+}
\end{equation}

The $max(.)$ operator ensures that the variance increases. We refer to the regularized adaptation in \eqref{eq:coral+} as the CORAL+ algorithm, while \eqref{eq:coral+_wo_reg} corresponds to the CORAL+ algorithm without regularization. Algorithm \ref{alg:coral+} summarizes the CORAL+ algorithm. Figure \ref{fig:regul} shows a plot of the diagonal elements of the term $(\mathbf{E}_{\rm b}$ - $\mathbf{I})$ in \eqref{eq:coral+}. Those entries with negative variances were removed automatically by the $max(.)$ operator. It ensures that the uncertainty increases (or stays the same) in the adaptation process.

It is worth noticing that, one could recover the subspace matrices $\{\mathbf{F}, \mathbf{G}\}$ via EVD. Nevertheless, this is not generally required as scores could be computed by plugging in the adapted covariance matrices $\mathbf{\Phi}^{+}_{\rm b}$, $\mathbf{\Phi}^{+}_{\rm w}$ and $\mathbf{C}^{+} = \mathbf{\Phi}^{+}_{\rm b} + \mathbf{\Phi}^{+}_{\rm w}$ into \eqref{eq:plda_marginal} and \eqref{eq:plda_joint}.    
\begin{algorithm}[t!]
    \caption{The CORAL+ algorithm for unsupervised adaptation of PLDA.}
	\centering
	\includegraphics[width=3.2in]{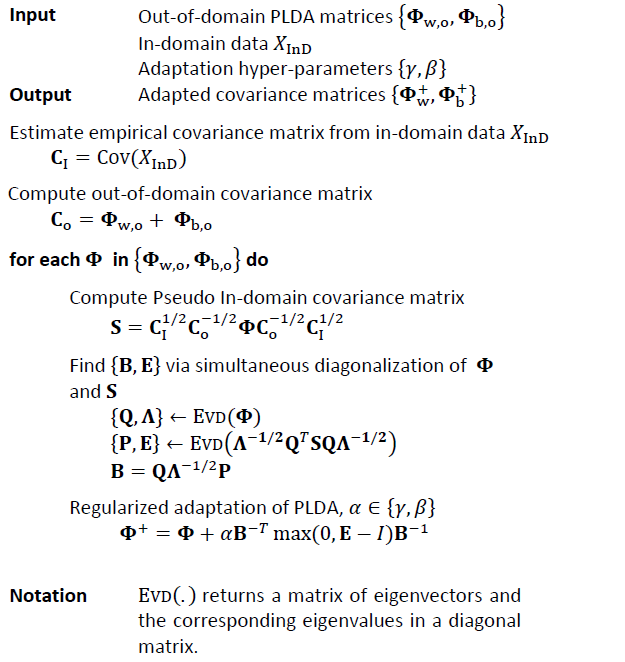}
	\label{alg:coral+}
	\vspace{-3ex}
\end{algorithm}

\section{Experiment}
Experiments were conducted on the the recent SRE'16 and SRE'18 datasets. The performance was measured in terms of \textit{equal error rate} (EER) and \textit{minimum detection cost} (MinCost)~\cite{sre16,sre18}. The latest SREs organized by NIST have been focusing on domain mismatch as one of the technical challenges. In both SRE'16 and SRE'18, the training set consists primarily English speech corpora collected over multiple years in the North America. This dataset encompasses Switchboard, Fisher, and the MIXER corpora used in SREs 04 -- 06, 08, 10, and 12. The enrollment and test segments are in Tagalog and Cantonese for SRE'16, and Tunisian Arabic for SRE'18. Domain adaptation was performed using the unlabeled subsets provided for the evaluation.       

The enrollment utterances have a nominal duration of 60 seconds, while the test duration ranges from 10 to 60 seconds. We used x-vector speaker embedding, which has shown to be very effective for speaker verification task over short utterances. (Recent results show that i-vector is more effective for longer utterance of over 2 minutes). The x-vector extractor follows the same configuration and was trained using the same setup as the Kaldi recipe~\footnote{https://github.com/kaldi-asr/kaldi/tree/master/egs/sre16/v2~\label{kaldi}}. A slight difference here is that we used an attention model in the pooling layer and extended the data augmentation~\cite{okabe2018attentive}. 

In our experiments, the dimension of the x-vector was 512. As commonly used in most state-of-the-art systems, LDA was used to reduce the dimensionality. We investigated the cases of 150- and 200-dimensional x-vector after LDA projection. CORAL~\cite{Sun2016} transformation was applied on the raw x-vectors before LDA. The transformed, and then projected x-vectors were used to train a PLDA for the {\tt CORAL PLDA} baseline. It is worth noticing that the LDA projection matrix was computed from the raw x-vectors, from which the CORAL transformation was also derived. We find that this gives the best performance compared to that reported in \cite{Alam2018}.

The proposed CORAL+ is a model-based adaptation technique. Domain adaptation is achieved by adapting the parameters (i.e., covariance matrices) pertaining to the {\tt OOD PLDA} as in \eqref{eq:coral+} and Algorithm \ref{alg:coral+} using the unlabeled in-domain dataset. The adaptation parameters were set empirically to $0.80$ in the experiments. Tables \ref{table:sre16} and \ref{table:sre18} show the performance of the baseline PLDA model trained on the out-of-domain English dataset ({\tt OOD PLDA}), the PLDA trained on the x-vectors which have been adapted using CORAL ({\tt CORAL PLDA}), and the OOD PLDA adapted to in-domain with CORAL+ algorithm ({\tt CORAL+ PLDA}). Also shown in the tables is the CORAL+ adaptation without regularization ({\tt w/o reg}). This correspond to the use of \eqref{eq:coral+_wo_reg} replacing \eqref{eq:coral+} in Algorithm \ref{alg:coral+}. 

The results on both SRE'16 and SRE'18 show consistent improvement of {\tt CORAL+ PLDA} compared to the {\tt OOD PLDA} baseline. The relative improvement amounts to $36.6\%$ and $22.35\%$ reduction in EER, and $32.0\%$ and $23.0\%$ reduction in MinCost on SRE'16 and SRE'18, respectively, for LDA dimension of $200$. Also shown in the tables is an unsupervised adaptation method implemented in Kaldi~\textsuperscript{\ref{kaldi}} ({\tt Kaldi PLDA}). The proposed {\tt CORAL+ PLDA} consistently outperforms this baseline on both SRE'16 and SRE'18 though the improvement over this baseline is more apparent on SRE'18. At LDA dimension of 200, the relative improvement amounts to $10.5\%$ reduction in EER, and $6.0\%$ reduction in MinCost on SRE'18. 

Compared to the feature-based CORAL ({\tt CORAL PLDA}), the benefit of CORAL+ ({\tt CORAL+ PLDA}) is more apparent on SRE'18. We obtained a relative reduction of $9.7\%$ in EER and $9.1\%$ in MinCost at LDA dimension of $200$. 
It is worth mentioning that SRE'16 has a unlabeled set with about the same size compared to that of SRE'18. Nevertheless, SRE'18 unlabeled set exhibits less variability (speaker and channel). This also explains the benefit of regularized adaptation on SRE'18 when a smaller and constrained unlabelled dataset is available for domain adaptation.       
\vspace{-2ex}
\section{Conclusion}
\vspace{-1ex}
We have presented the CORAL+ algorithm for unsupervised adaptation of PLDA backend to deal with the domain mismatch issue in practical applications. Similar to the feature-based correlation alignment (CORAL) technique, the CORAL+ domain adaptation is accomplished by matching the out-of-domain statistics to that of the in-domain. We show that statistics matching could be directly applied on PLDA model. We further improve the robustness by introducing additional adaptation parameter and regularization to the adaptation equation. The proposed method shows significant improvement compared to the PLDA baseline. Results also show the benefit of model-based adaptation especially when the data available for adaptation is relatively small and constrained.

\begin{table}[t!]
  \small
  	\caption{{\it Performance comparison on SRE'16 (CMN). The dimension of x-vector after LDA is $150$ and $200$. Boldface denotes the best performance for each column.}} 
  	\vspace{-4ex}
  	\begin{center}
  		\begin{tabular}{l|c|c|c|c}
  			\hline
  			                            &  \multicolumn{2}{c|}{LDA $150$}  & \multicolumn{2}{c}{LDA $200$}      \\ \hline  
			                            & EER (\%) 	& MinCost              & EER (\%) 	    & MinCost           \\ \hline
  			{\tt OOD PLDA} 			    & 9.69  	& 0.783		           & 9.94  	        & 0.813             \\ 
  			{\tt Kaldi PLDA}		    & 6.82 	    & 0.552		           & 6.57  	        & 0.558	            \\
  			{\tt CORAL PLDA}		    & \bf{6.50} & \bf{0.539}           & 6.31  	        & \bf{0.543}        \\
  			{\tt CORAL+ PLDA} 		    & 6.62      & 0.540		           & \bf{6.30}	    & 0.553             \\
  			{\tt w/o reg} 	            & 6.93  	& 0.544		           & 6.51  	        & 0.547        \\ \hline			
   		\end{tabular}
  	\end{center}
  	\label{table:sre16}
  	\vspace{-3ex}
\end{table} 
\begin{table}[t!]
  \small
  	\caption{{\it Performance comparison on SRE'18 (CMN2). The dimension of x-vector after LDA is $150$ and $200$. Boldface denotes the best performance for each column.}}
  	\vspace{-4ex}
  	\begin{center}
  		\begin{tabular}{l|c|c|c|c}
  			\hline
  			                            &  \multicolumn{2}{c|}{LDA $150$}  & \multicolumn{2}{c}{LDA $200$}          \\ \hline  
			                            & EER (\%) 	& MinCost               & EER (\%) 	    & MinCost               \\ \hline
  			{\tt OOD PLDA} 			    & 7.19  	& 0.538  		        & 7.47  	    & 0.569                 \\ 
  			{\tt Kaldi PLDA}		   	& 6.25  	& 0.435                 & 6.48  	    & 0.466                 \\
  			{\tt CORAL PLDA}		   	& 6.22 		& 0.449                 & 6.42  	    & 0.482                 \\
  			{\tt CORAL+ PLDA} 		   	& \bf{5.95} & \bf{0.421}            & \bf{5.80}	    & \bf{0.438}            \\
  			{\tt w/o reg} 	            & 6.49 	    & 0.441                 & 6.33  	    & 0.460                 \\ \hline			
        \end{tabular}
  	\end{center}
  	\label{table:sre18}
  \end{table}

\vfill\pagebreak


\bibliographystyle{IEEEbib}
\bibliography{refs}

\end{document}